\documentclass[10pt,twocolumn,letterpaper]{article}

\usepackage[pagenumbers]{cvpr}

\usepackage{graphicx}
\usepackage{algorithm}
\usepackage{algorithmic}
\usepackage{amsmath}
\usepackage{booktabs}
\usepackage{amssymb}
\usepackage{mathtools}
\usepackage{xcolor} 
\usepackage{multirow}

\definecolor{cvprblue}{rgb}{0.21,0.49,0.74}
\usepackage[pagebackref,breaklinks,colorlinks,allcolors=cvprblue]{hyperref}

\title{Generative Semantic Coding for Ultra-Low Bitrate Visual Communication and Analysis}

\author{Weiming Chen$^1$\thanks{Equal contributions} \quad Yijia Wang$^{1*}$ \quad Zhihan Zhu$^1$ \quad Zhihai He$^{1,2}$\thanks{Corresponding author}\\
$^1$Southern University of Science and Technology, Shenzhen, China\\
$^2$Pengcheng Laboratory, Shenzhen, China\\
{\tt\small \{chenwm2023,wangyj2022,12312326\}@mail.sustech.edu.cn}\\
{\tt\small hezh@sustech.edu.cn}
}

\begin{document}
\maketitle

\begin{abstract}
We consider the problem of ultra-low bit rate visual communication for remote vision analysis, human interactions and control in challenging scenarios with very low communication bandwidth, such as deep space exploration, battlefield intelligence, and robot navigation in complex environments. In this paper, we ask the following important question: \textbf{can we accurately reconstruct the visual scene using only a very small portion of the bit rate in existing coding methods while not sacrificing the accuracy of vision analysis and performance of human interactions?} Existing text-to-image generation models offer a new approach for ultra-low bitrate image description. However, they can only achieve a semantic-level approximation of the visual scene, which is far insufficient for the purpose of visual communication and remote vision analysis and human interactions. To address this important issue, we propose to seamlessly integrate image generation with deep image compression, using joint text and coding latent to guide the rectified flow models for precise generation of the visual scene. The semantic text description and coding latent are both encoded and transmitted to the decoder at a very small bit rate. Experimental results demonstrate that our method can achieve the same image reconstruction quality and vision analysis accuracy as existing methods while using much less bandwidth. The code will be released upon paper acceptance.
\end{abstract}

\section{Introduction}
\label{Introduction}

In this paper, we consider the problem of ultra-low bit rate visual communication for remote vision analysis, human interactions and control in challenging scenarios such as deep space exploration, battlefield intelligence, and robot navigation in complex environments.
In these scenarios, the sender and the receiver often have abundant computational power and resources. For example, the exploration robot on the moon or Mars, as well as the receiving station, is equipped with high-end GPUs and a sufficient power supply. However, the communication bandwidth between the sender and receiver is a very scarce resource due to the long transmission distance or strong interference.
In these scenarios, we need to accurately reconstruct the visual scene for vision analysis, decision making, human interactions and control. 

\begin{figure}
    \centering
    \includegraphics[width=\linewidth]{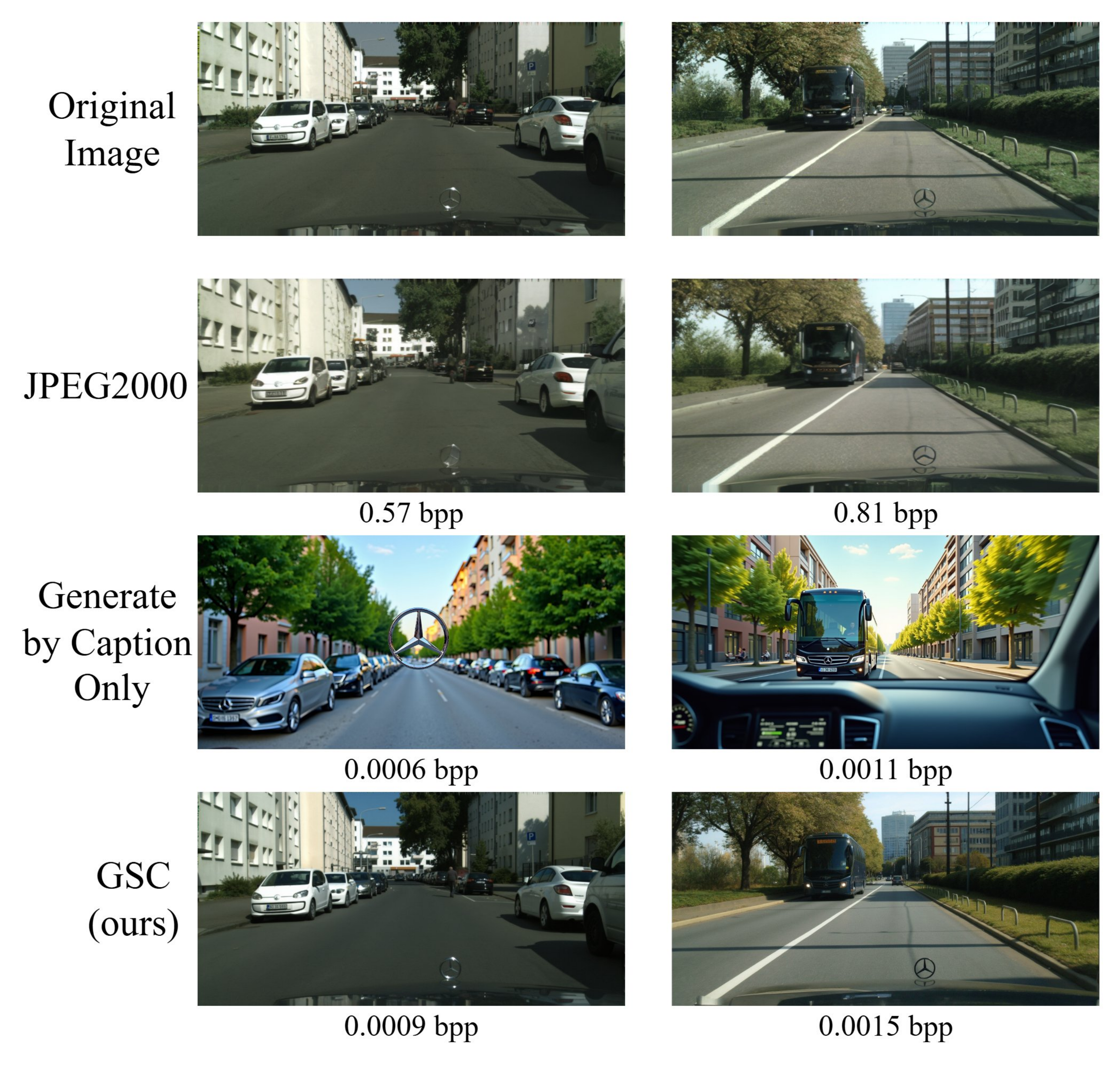}
    \caption{Example of GSC result compared with JPEG2000, one of the compression standards, and the result generated only guided by the caption.}
    \label{fig:enter-label}
\end{figure}

\begin{figure*}
    \centering
    \includegraphics[width=\textwidth]{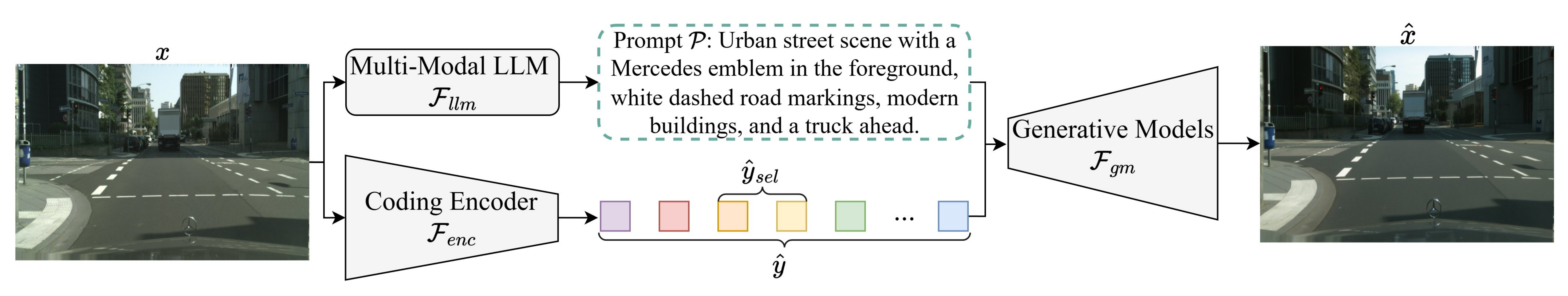}
    \caption{Overview of the proposed Generative Semantic Coding (GSC) framework.}
    \label{fig: framework}
\end{figure*}

Existing image and video compression methods, such as JPEG2000~\cite{ISO_IEC_15444_1_2000} and H.265~\cite{h265standard} excel in pixel-level reconstruction, but require high bandwidth. For instance, H.265 encoded standard-definition video often requires bandwidth ranging from 1 Mbps to 2 Mbps, which far exceeds the bandwidth available in these scenarios discussed above.
Note that, in these scenarios, the purpose of the visual communication is to support accurate remote vision analysis, human interactions, control, and decisions. 
Thus, pixel-perfect reconstruction is not necessary. It only needs to reconstruct the image and visual scene such that subsequent vision analysis performance is consistent with that using the original images. 
An important question to ask is: 
\textit{can we accurately reconstruct the visual scene using only a very small portion of the bit rate in existing coding methods while not sacrificing the accuracy of vision analysis and performance of human interactions?}

Recent advances in text-to-image generation models~\cite{qiao2019mirrorgan, ramesh2021zero, ding2021cogview, goodfellow2014generative, goodfellow2020generative, rombach2022high, ramesh2022hierarchical, saharia2022photorealistic, esser2024scaling} offer a novel approach 
for scene description and reconstruction. With this method, we only need to transmit the text descriptions of the scene to the receiver end, allowing the reconstruction of the visual scene. Unfortunately, 
the text description is often very subjective. With texts, they can only reconstruct and approximate the visual scene semantically at a very coarse level. 
Recently, researchers have studied using extra visual information, such as contours and sketches, to guide the text-image generation process  ~\cite{rombach2022high, theis2022lossy, lei2023text+}. They still suffer from inaccurate reconstruction of image details and high bit rate cost.

To overcome these limitations, we proposed a novel framework, called Generative Semantic Coding (GSC), as illustrated in Figure \ref{fig: framework}.  We seamlessly integrate image generation with deep image compression, using joint text and coding latent to guide the rectified flow models for precise generation of the visual scene. We observe that the coded latents from the deep image compression system provide compact and high-quality guidance for the image generation. 
We dynamically select a tiny portion of the coding latents that contains the most significant information for preserving the structural consistency between the original and reconstructed images. 
The semantic text description and coding latent are both encoded and transmitted to the decoder at a very small bit rate. As shown in Figure~\ref{fig:enter-label}, these selected coding latents only require less than 0.001 bpp, which is ultra-low but works well.

\section{Related Work and Unique Contribution}
\label{RelatedWork}

In this section, we first review existing generative image compression methods related to our work. Then, we point out the necessity of adding conditional guidance. Finally, we summarize the unique contributions of this work.

\subsection{Ultra-Low Bitrate Coding with Generative Models}
\label{sec: Limitations related work}
Existing generative image compression methods typically for ultra-low bitrates operate within the bitrate range from 0.02 bpp to 0.10 bpp. For example, GLC~\cite{jia2024generative} and HiFiC~\cite{mentzer2020high} employ GANs to learn image distributions for efficient compression but suffer from significant distortions and detail loss at extremely low bitrates. PerCo~\cite{careil2024towards} trains a hyper-encoder and a codebook to extract image features, emphasizing perceptual quality via diffusion models; nevertheless, at extremely low bitrates, its perceptual quality still degrades. MS-ILLM~\cite{muckley2023improving} optimizes compression through multi-step iterations and language models to extract semantic information, but its image quality is severely compromised below 0.01 bpp. Recently, some methods~\cite{yang2023lossy, pan2022extreme} transmit a quantized embedding as a conditional input to the diffusion-based decoder, while DiffC~\cite{theis2022lossy, vonderfecht2025lossy} directly transmits pixels corrupted by noise in a diffusion process. But they don't focus on the semantic coding.
Text-Sketch~\cite{lei2023text+} adopts prompt inversion to maintain semantic consistency through CLIP~\cite{radford2021learning}, but struggles to keep spatial consistency and wastes lots of bits. These methods all face challenges at bitrates lower than 0.01 bpp, highlighting the need for more advanced techniques to address this issue. We leverage the inferent structural information embedded in the coded feature to ensure consistency under extremely low bitrate conditions.
 
\subsection{Controllable Diffusion Models}
\label{sec: conditional guidance related work}
One limitation of generative image compression methods is that textual descriptions alone cannot effectively control the image generation process. Therefore, it is necessary to incorporate additional conditional guidance mechanisms~\cite{zhang2023adding,ye2023ip,mou2024t2i} to enhance controllability. ControlNet~\cite{zhang2023adding} augments diffusion models with additional conditional branches, enabling fine-grained control over the generation process using structural information, while preserving the original model's generation fidelity. IP-Adapter~\cite{ye2023ip} introduces a decoupled cross-attention mechanism by adding an additional cross-attention module to each existing cross-attention layer in the U-Net, facilitating more effective identity or style transfer in text-to-image generation. T2I-Adapter~\cite{mou2024t2i} introduces lightweight and composable adapters that align internal features of frozen text-to-image models with external control information. Inspired by ControlNet~\cite{zhang2023adding}, we augment the FLUX model~\cite{flux2024} with an additional module to inject encoded guidance, effectively controlling image generation and preserving both structural and semantic information.

\subsection{Unique contributions}
\label{subsec: contributions related work}
Our major unique contributions are as follows: (1) This paper considers an extreme scenario where transmission resources are severely limited while side resources are abundant. In this context, we discuss how to encode an image using minimal information, targeting bitrates below 0.01 bpp. (2) We develop a new approach, called generative semantic coding (GSC), which controls the image generation process to reconstruct images as precisely as possible. (3) Extensive experiments on three fundamental vision tasks demonstrate that our method achieves comparable performance to previous approaches while only utilizing less than $\mathrm{10\%}$ of their bpp, specifically less than 0.007 bpp.

\section{The Proposed GSC Method}
\label{Method}

In this section, we begin with an overview of our proposed method (Section~\ref{subsec: method overview}), followed by a detailed exposition of its two principal components (Section~\ref{subsec: latent construction and channel selection} and Section~\ref{subsec: image reconstruction}). Finally, we provide a theoretical analysis of the problem and our method (Section~\ref{subsec: theoretical analysis}).

\subsection{Method Overview}
\label{subsec: method overview}
The architecture of our proposed GSC framework is shown in Figure~\ref{fig: framework}. Given an input image $x$, we first extract its caption $\mathcal{P}$ by a multi-model large language model (MM-LLM). This caption encodes the semantic information of $x$. Structural and spatial details are extracted by a deep image encoder $\mathcal{F}_{enc}$ that generates the latent representation $\hat{y}=\{\hat{Y}_1, \hat{Y}_2, \hat{Y}_3, \dots, \hat{Y}_{n}\}$, from which, we dynamically select a small subset  $\hat{y}_{sel} = \{\hat{Y}^{sel}_1, \hat{Y}^{sel}_2, \hat{Y}^{sel}_3, \dots, \hat{Y}^{sel}_{C} \}$. Both $\mathcal{P}$ and $\hat{y}_{sel}$ are encoded and transmitted to the receiver. Guided by the $\mathcal{P}$ and $\hat{y}_{sel}$, the receiver reconstructed image $\hat{x}$ by the rectified flow (RF)~\cite{liu2022flow}. $P$ enforces the semantic consistency between the original and reconstructed image, while  $\hat{y}_{sel}$ ensures the structural consistency.

\subsection{Latent Construction and Channel Selection}
\label{subsec: latent construction and channel selection}

As stated in the above section, we obtain the latent representation $\hat{y}$ by encoding the original image $x$ with a pre-trained image coding encoder. Instead of transmitting all $n$ channels of $\hat{y}$, we focus on selecting $C$ channels of $\hat{y}$ to guide the generation process. Here, $C$ is a very small number. 

We first use the deep image encoder $g_a$ to analyze the input image $x$  to generate the latent representation  $y = g_a \left ( x ; \phi \right )$, where $\phi$ is the learned parameters of $g_a$. Then, $y$ is quantized into $\hat{y}$ using a quantizer $Q$, $\hat{y} = Q \left ( y \right )$. The entropy model is used to estimate the probability distribution $\Phi$ of $\hat{y}$ to optimize bit allocation in encoding and decoding processes.  This process can be written as:
\begin{equation}
    \hat{y} = Q \left ( g_a \left (x, \phi \right), \Phi \right ).
\end{equation}

\begin{figure}
    \centering
    \includegraphics[width=\linewidth]{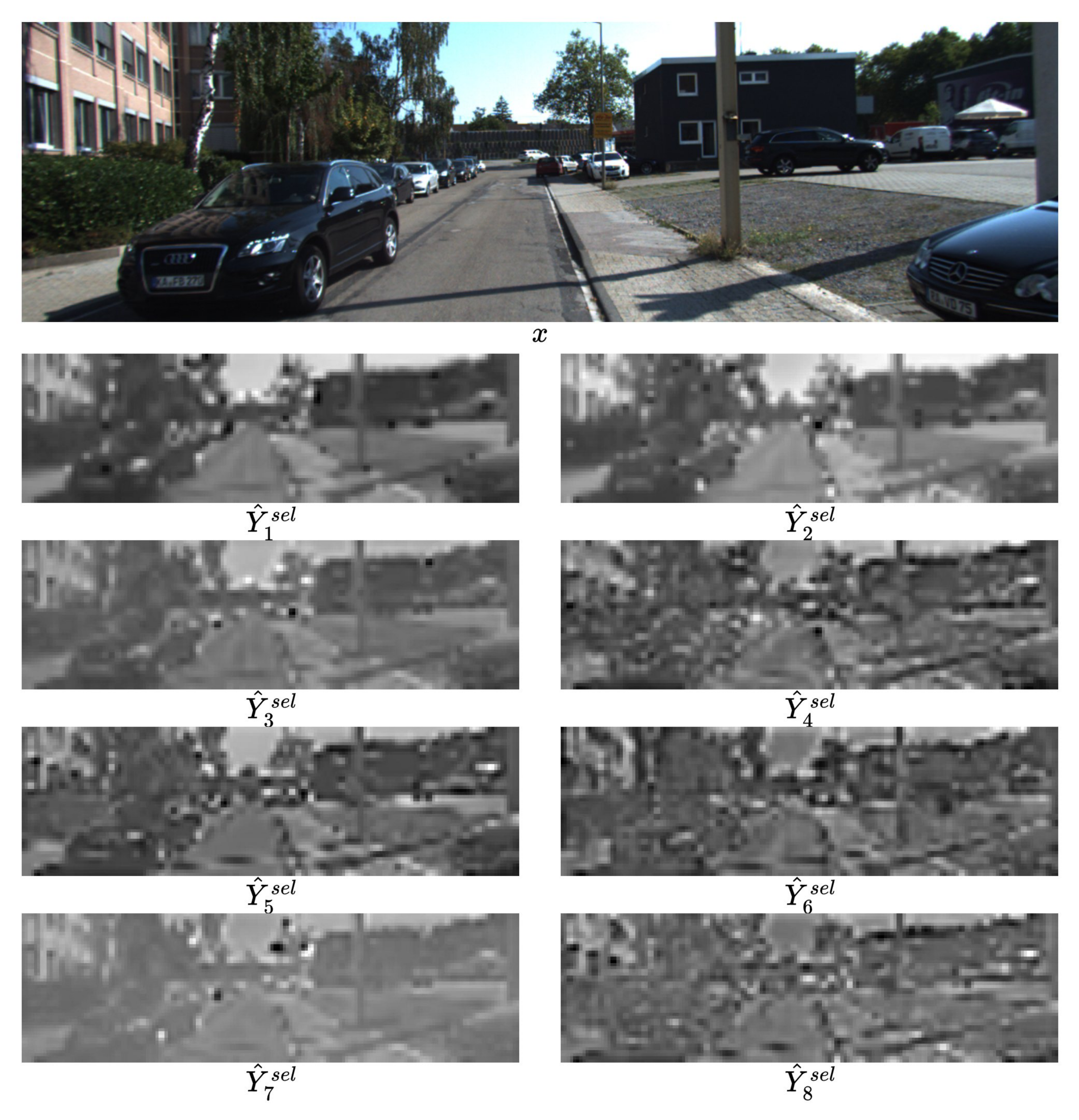}
    \caption{An example visualization of 8 selected channels.}
    \label{fig: channels}
\end{figure}

From $\hat{y}$, we select a very small subset of channels  $\hat{y}_{sel}$ to guide the image generation process.
In this work, we recognize that the task of  $\hat{y}_{sel}$ is to maintain the spatial and structural consistency between the reconstructed image and the original input. Therefore, we propose to use the SSIM (Structural Similarity Index) to dynamically select $\hat{y}_{sel}$. 
In our design, we select $C$ channels with the largest SSIM value computed from the $\hat{Y}_{i} \left ( i = 1, 2, 3, \dots, 320 \right )$ and an example of the gray-scale representation of the selected channels, i.e., $\hat{Y}_{i}^{sel} \left( i = 1, 2, 3, \dots, 8 \right )$, is shown in Figure~\ref{fig: channels}.

It should be noted that,  if more channels are selected to construct the $\hat{y}_{sel}$, higher accuracy can be achieved; however, more bits are required to encode them. The represents  a tradeoff between the visual analysis performance and encoding bit rate
\begin{equation}
    \min_{\hat{y}_{sel}} \alpha \left| V(\hat{x}) - V(x) \right| + \beta B(\hat{y}_{sel}, \mathcal{P}),
\end{equation}
where $B(\hat{y}_{sel}, \mathcal{P})$ represents the bits required to transmit the $\hat{y}_{sel}$ and $\mathcal{P}$, $V(x)$ represents the visual analysis results of $x$, and $\alpha$ and $\beta$ are weight parameters to control the trade-off between them.

\subsection{Joint Text-Latent Guided Image Generation}
\label{subsec: image reconstruction}

\begin{figure}
    \centering
    \includegraphics[width=\linewidth]{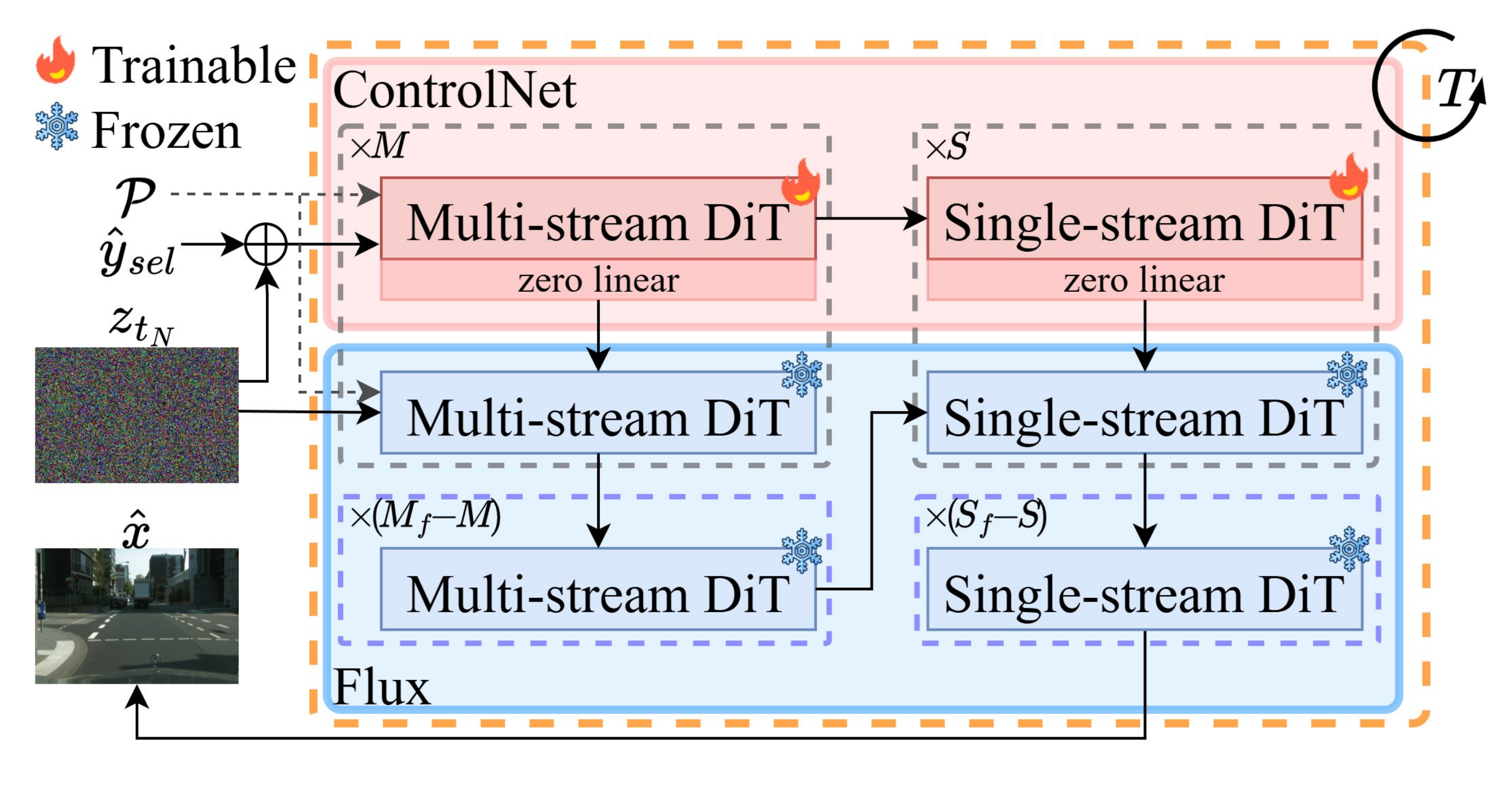}
    \caption{The details of the text-structural image generation process.}
    \label{fig: flux}
\end{figure}

As stated in the above section, guided by the image description $\mathcal{P}$ and its coding latent $\hat{y}_{sel}$, we generate the reconstructed image $\hat{x}$ using the FLUX text-to-image generation model~\cite{flux2024}. 
As shown in Figure~\ref{fig: flux}, a noise latent $z_{t_{N}}$ is randomly sampled from the Gaussian distribution $\mathcal{N} \left ( 0, \mathbf{I} \right )$. It is denoised under the guidance of $\mathcal{P}$ and $\hat{y}_{sel}$. The $\mathcal{P}$ is directly input into the T5 text encoder~\cite{Raffel2020T5} to become the text embedding $\mathcal{P}_{emb}$ to be used in the following Diffusion Transformer (DiT)~\cite{peebles2023scalable} blocks. To incorporate the guidance of $\hat{y}_{sel}$, we create a trainable copy of the $M$ multi-stream DiT blocks and $S$ single-stream DiT blocks. Its initial inputs contain two parts: one is $\mathcal{P}_{emb}$, and the other is the sum of $z_{t_{N}}$ and $\hat{y}_{sel}$. The outputs of each corresponding DiT block, after passing through the zero linear layer, are added to the first $M$ multi-stream DiT blocks and $S$ single-stream DiT blocks, respectively. As there are $M_f$ Multi-stream DiT blocks and $M_s$ Single-stream DiT blocks in the original FLUX~\cite{flux2024}, the rest $\left ( M_f - M \right )$ multi-stream DiT blocks and $\left ( S_f - S \right )$ single-stream DiT blocks remain the same as the original ones. After that, it performs denoising over $N$ discrete timesteps $t = \left \{ t_{N}, \dots, t_{0} \right \}$ by the following equation:
\begin{equation}
\begin{split}
z_{t_{i - 1}} = z_{t_{i}} + \left( t_{i - 1} - t_{i} \right) v_{\theta} \left( z_{t_{i}}, t_{i}, \mathcal{P}_{\text{emb}}, \hat{y}_{\text{sel}} \right), 
\end{split}
\end{equation}
where $i = N, N-1, N - 2, \dots, 1$ and $v_{\theta}$ is the predicted vector field obtained from the DiT blocks, parameterized by $\theta$. After $z_0$ is obtained, it serves as an input to the VAE decoder to obtain the final output image. After $T$ steps, we finally obtain the $\hat{x}$.

For training, we only activate and train $M$ multi-stream DiT blocks and $S$ single-stream DiT blocks, and freeze all the DiT blocks in the original FLUX~\cite{flux2024}. The goal is to train a neural network to predict the $v_{\theta}$. To this end, we couple samples from the target distribution with the samples from the Gaussian distribution via a linear path: $Z_t = tZ_1 + (1-t)Z_0$. Therefore, the marginal distribution of $Z_t$ becomes: 
\begin{equation}
p_t({z}_t) = \mathbb{E}_{Z_1 \sim p_1} \left[ p_t(z_t | Z_1) \right] = \int p_t(z_t | z_1) p_1(z_1) \, dz_1.
\end{equation}
Given the initial state $Z_0=z_0$ and the target state $Z_1=z_1$, the linear path becomes $dZ_t = v_t(Z_t|z_1)dt = z_1 - z_0$. The marginal vector field can be derived from the conditional vector field using the equation as follows,
\begin{equation}
\begin{split}
    v_t(z_t) = \mathbb{E}_ {Z_1 \sim p_1} \left[ v_t(z_t|Z_1) \frac{p_t(z_t|Z_1)}{p_t(z_t)} \right] \\
    = \int v_t(z_t|z_1) \frac{p_t(z_t|z_1)}{p_t(z_t)} p_1(z_1) dz_1.
\end{split}
\end{equation}
After that, we use a neural network $v_{\theta}(z_t, t, \mathcal{P}, \hat{y}_{sel})$, parameterized by $\theta$, to approximate the marginal vector field $v_t(z_t)$ through the conditional flow matching given by
\begin{equation}
\begin{split}
    \mathcal{L}_{CFM}(\varphi) \coloneqq \mathbb{E}_{t \sim \mathcal{U}[0,1],\, Z_t \sim p_t(\cdot \,|\, Z_1),\, Z_1 \sim p_1} \\
    \left[ \| v_t(Z_t|Z_1) - v_{\theta}(Z_t, t, \mathcal{P}_{emb}, \hat{y}_{sel}; \varphi) \|_2^2 \right].
\end{split}
\end{equation}

\begin{table*}
    \centering
    \caption{Depth estimation results on KITTI and Hypersim.}
    \resizebox{\linewidth}{!}{
    \begin{tabular}{lcccccccccccccc}
        \toprule
        \multirow{2}{*}{\textbf{Method}} & \multicolumn{7}{c}{KITTI} & \multicolumn{7}{c}{Hypersim} \\
        \cmidrule(lr){2-8} \cmidrule(l){9-15}
        & bpp$\downarrow$ & $\delta_1$$\uparrow$ & $\delta_2$$\uparrow$& $\delta_3$$\uparrow$ & AbsRel$\downarrow$ & RMSE$\downarrow$ & RMSE log$\downarrow$ & bpp$\downarrow$ & $\delta_1$$\uparrow$ & $\delta_2$$\uparrow$& $\delta_3$$\uparrow$ & AbsRel$\downarrow$ & RMSE$\downarrow$ & RMSE log$\downarrow$ \\
        \midrule
        Original & - & 0.891 & 0.981 & 0.995 & 0.109 & 3.557 & 0.149 & - & 0.746 & 0.937 & 0.978 & 0.184 & 1.281 & 0.216 \\
        Directly Gen. & 0.0063 & 0.255 & 0.538 & 0.720 & 0.756 & 15.906 & 0.611 & \textbf{0.0023} & 0.364 & 0.604 & 0.771 & 0.697 & 2.795 & 0.574\\
        PIC & \textbf{0.0011} & 0.381 & 0.628 & 0.773 & 0.644 & 14.371 & 0.581 & 0.0027 & 0.315 & 0.552 & 0.726 & 0.697 & 3.035 & 0.623 \\
        PICS & 0.0235 & 0.703 & 0.877 & 0.948 & 0.208 & 7.263 & 0.292  & 0.0259 & 0.554 & 0.786 & 0.990 & 0.302 & 2.009 & 0.390 \\
        PerCo\textsubscript{19} & 0.0037 & 0.619 & 0.816 & 0.910 & 0.300 & 8.631 & 0.361 & 0.0037 & 0.440 & 0.693 & 0.810 & 0.465 & 2.317 & 0.492\\
        PerCo\textsubscript{313} & 0.0329 & 0.808 & 0.943 & 0.979 & 0.153 & 5.468 & 0.215 & 0.0329 & 0.691 & 0.893 & 0.940 & 0.233 & 1.499 & 0.275\\
        MS-ILLM\textsubscript{20} & 0.0079 & 0.197 & 0.352 & 0.493 & 0.483 & 14.108 & 0.953  & 0.0049 & 0.276 & 0.501 & 0.663 & 0.484 & 2.936 & 0.674\\
        MS-ILLM\textsubscript{40} & 0.0124 & 0.506 & 0.699 & 0.799 & 0.283 & 10.594 & 0.529 & 0.0075 & 0.438 & 0.688 & 0.808 & 0.379 & 2.322 & 0.480\\
        MS-ILLM\textsubscript{350} & 0.0539 & 0.777 & 0.912 & 0.959 & 0.149 & 6.519 & 0.255 & 0.0327 & 0.650 & 0.891 & 0.944 & 0.223 & 1.522 & 0.270\\
        \midrule
        Ours ($C=1$)& 0.0069 & 0.796 & 0.931 & 0.976 & 0.170 & 5.652 & 0.223  & 0.0026 & 0.615 & 0.837 & 0.907 & 0.288 & 1.763 & 0.334 \\
        Ours ($C=2$) & 0.0074 & 0.834 & 0.950 & 0.985 & 0.148 & 4.947 & 0.197 & 0.0028 & 0.627 & 0.847 & 0.907 & 0.254 & 1.667 & 0.304 \\
        Ours ($C=4$) & 0.0081 & 0.852 & 0.965 & 0.989 & 0.134 & 4.495 & 0.179 & 0.0032 & 0.663 & 0.896 & 0.934 & 0.227 & 1.492 & 0.269\\
        Ours ($C=8$) & 0.0104 & 0.859 & 0.968 & 0.991 & 0.129 & 4.386 & 0.174 & 0.0043 & 0.668 & 0.892 & 0.952 & 0.225 & 1.447 & 0.261\\
        Ours ($C=16$) & 0.0150 & \textbf{0.866} & \textbf{0.973} & \textbf{0.992} & \textbf{0.123} & \textbf{4.051} & \textbf{0.166} & 0.0064 & \textbf{0.711} & \textbf{0.912} & \textbf{0.959} & \textbf{0.211} & \textbf{1.375} & \textbf{0.242}\\
        \bottomrule
    \end{tabular}    
    }
    \label{tab: depth}
\end{table*}

\subsection{Theoretical Analysis}
\label{subsec: theoretical analysis}
In image compression with text and structural information, some guidance information might be useless or even misleading for the target image generation process. For example, different channels of $\hat{y}_{sel}$ might contain similar information. Although it is difficult to accurately extract the useful guidance information, it is very important to understand its performance bound. Here, we present a theoretical analysis to characterize the lower bound of the coding bit rate.

We recognize that useful information is not uniformly distributed throughout the entire image, and only a subset of pixels contains important and useful information about the image. Motivated by this, we introduce a function $U(X)$ to quantize the information contained by pixel $X$ in the image $x$ has. We obtain the probability of quantized information by 
\begin{equation}
    P(X) = \frac{U(X)}{\sum_{X_i \in x} U(X_i)}, \quad P(E|X) = \frac{P(E \cap x)}{P(x)}
\end{equation}
where $E$ represents the information in the image $x$. The information entropy by the given image $x$ is $H(E|x) = - \sum_{X_i \in x} P (E|x) \cdot logP(E|x)$. As more proper vision analysis information an image contains, the larger its entropy value will be. So, $V(x) \propto H(x)$. According to the rate-distortion function~\cite{shannon1959coding}, the compression rate $R=B(\hat{y}_{sel}, \mathcal{P})$ should be no less than the entropy of $\hat{x}$. Therefore, it can be formulated as the following optimization problem:
\begin{equation}
\begin{aligned}
\min_{\hat{x}, ~\hat{y}_{sel}} \quad & \alpha ( H(x) - H(\hat{x})) + \beta B(\hat{y}_{sel}, \mathcal{P}),\\
\text{s.t.} \quad & R \geq H ( \hat{x}).
\end{aligned}
\end{equation}
As $\hat{x}$ is obtained from denoising a sample from $\mathcal{N} ( 0, \mathbf{I})$, it follows with a normal distribution $\mathcal{N}(\mu, \Sigma)$. So, we can use the Lagrange multiplier method to find the solution even though it is not a convex problem. The constructed Lagrange function is,
\begin{equation}
\begin{split}
        L ( \hat{x}, \hat{y}_{sel}, \lambda) = \alpha ( H(x) - H(\hat{x})) + \beta B(\hat{y}_{sel}, \mathcal{P}) \\ 
        + \lambda ( H(\hat{x}) - R).
\end{split}
\end{equation}
And the theoretical optimal solution occurs when
\begin{equation}
    \nabla_{\hat{x}, ~\hat{y}_{sel}, ~\lambda} L (\hat{x}, \hat{y}_{sel}, \lambda ) = 0.
\end{equation}

\section{Experimental Results}
\label{sec: Experiment}

In this section, we provide extensive experimental results to evaluate the proposed GSC method and ablation studies to understand its performance and evaluate its robustness.

\begin{figure}[t]
    \centering
    \includegraphics[width=\linewidth]{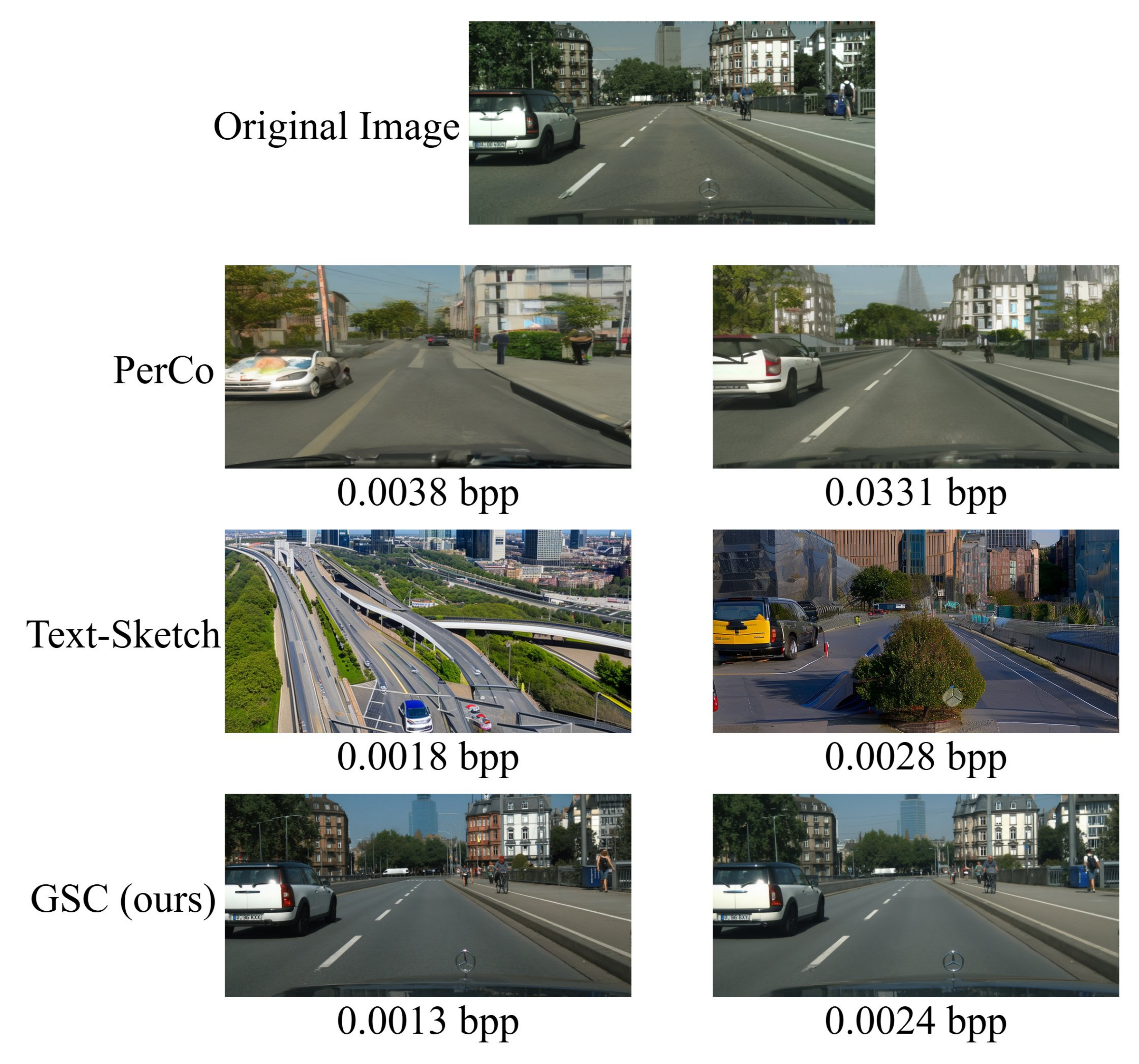}
    \caption{Qualitative results of CityScapes compared with other methods.}
    \label{fig: qualitative results}
\end{figure}

\begin{figure*}[t]
    \centering
    \includegraphics[width=0.9\linewidth]{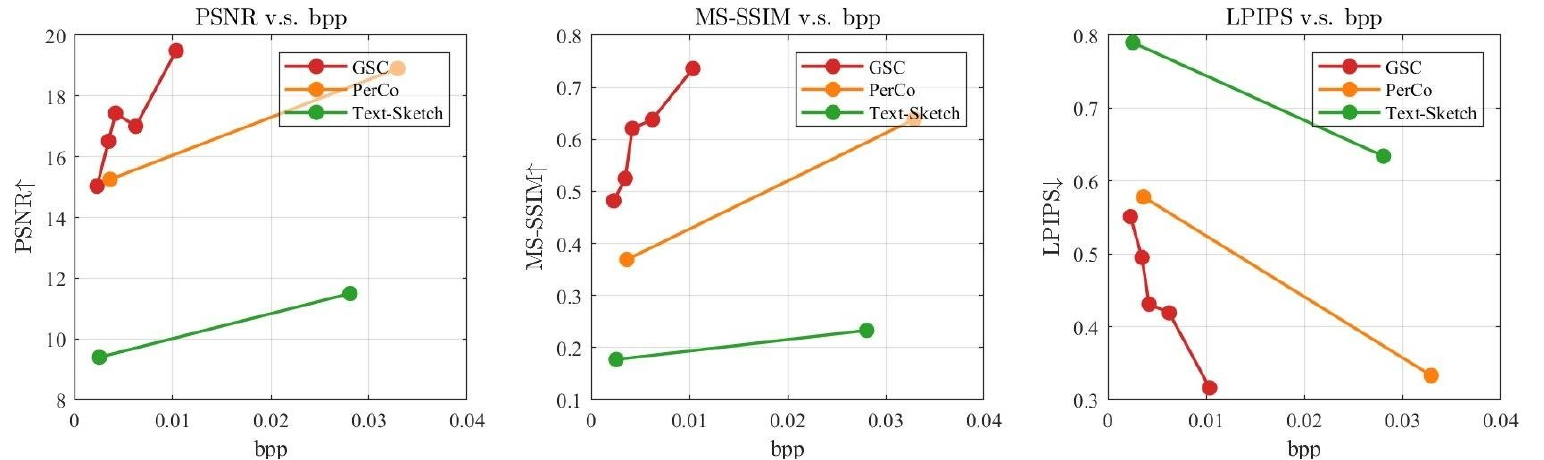}
    \caption{The rate-distortion performance comparison of different methods on the Kodak dataset.}
    \label{fig: rd curve of Kodak}
\end{figure*}

\begin{table}[t]
    \centering
    \caption{Pixel-level semantic segmentation result on the subset of CityScapes.}
    \label{tab: pix-semantic-seg}
    \resizebox{0.8\linewidth}{!}{
    \begin{tabular}{lcccc}
        \toprule
        Method & bpp$\downarrow$ & aAcc$\uparrow$ & mIoU$\uparrow$ & mAcc$\uparrow$ \\
        \midrule
        Original & - & 96.370 & 82.280 & 88.600  \\
        Directly Gen. & \textbf{0.0009} & 39.430 & 8.450 & 15.280  \\
        PIC & 0.0020 & 36.200 & 6.450 & 11.860\\ 
        PICS & 0.0299 & 61.870 & 18.40 & 30.390\\
        PerCo\textsubscript{19} & 0.0037 & 67.260 & 21.700 & 31.170 \\
        PerCo\textsubscript{313} & 0.0330 & 82.470 & 38.300 & 47.520\\
        MS-ILLM\textsubscript{20} & 0.0038 & 63.530 & 15.610 & 22.110\\
        MS-ILLM\textsubscript{40} & 0.0064 & 76.180 & 29.380 & 38.690\\
        MS-ILLM\textsubscript{350} & 0.0310 & 90.600 & 60.680 & 70.860 \\
        \midrule
        Ours ($C=1$) & 0.0011 & 85.250 & 47.840 & 58.360 \\
        Ours ($C=2$) & 0.0013 & 88.370 & 54.300 & 65.060 \\
        Ours ($C=4$) & 0.0015 & 90.540 & 61.090 & 71.510  \\
        Ours ($C=8$) & 0.0023 & 90.940 & 63.440 & 71.570 \\
        Ours ($C=16$) & 0.0039 & \textbf{93.440} & \textbf{70.730} & \textbf{78.090} \\
        \bottomrule
        \end{tabular}
        }
\end{table}

\subsection{Experimental Settings}
\label{subsec: experimental settings}

\textbf{(1) Datasets.} For training the model, 20,000 images were constructed by randomly sampling 5,000 images from the training sets of KITTI~\cite{Uhrig2017THREEDV}, Flickr30k~\cite{flickr30k}, COCO2017~\cite{lin2014microsoft}, and iNaturalist~\cite{van2018inaturalist}, respectively, and combining them together. This enhances the diversity of the datasets and thus ensures the generalizability of the model.

\noindent \textbf{(2) Implementation details.}
Our model was implemented using PyTorch and trained on a single NVIDIA HGX H20-96G GPU. The number of multi-stream and single-stream DiT blocks are set to $M = 4$ and $S = 2$, respectively. We trained the model for 15,000 steps using the AdamW optimizer with the learning rate and weight decay set to $4 \times 10^{-5}$ and 0.01, respectively. The batch size is set to 1, and gradients are accumulated for 4 steps during the training. In our model, we trained 5 models with fixed channels of 1, 2, 4, 8, and 16. For getting textual descriptions of images, we use the Qwen2.5-vl-72b-Instruct~\cite{qwen25}.

\subsection{Performance Comparisons}
\label{subsec: comparison on basic vision tasks}

We compare our methods with other ultra-low bitrate methods, including Text-Sketch~\cite{lei2023text+}, Perco~\cite{careil2024towards}, and MS-ILLM~\cite{muckley2023improving}. These methods can still achieve the state-of-the-art (SOTA) when the bit rate is lower than 0.01 bpp. Since our work focuses on the semantic coding at scenarios such as deep space exploration, with bitrates lower than 0.01 bpp, we assess the quality of reconstructed images by their downstream performance on fundamental vision tasks, and therefore adopt task-specific datasets. Specifically, we conduct evaluations across three vision tasks: depth estimation, semantic segmentation, and object detection. The goal is to evaluate whether the reconstructed images maintain sufficient information well needed for accurate vision analysis. In the following tables, the ``Directly Gen.'' means the result directly generated by FLUX~\cite{flux2024} using only the prompt to guide the generation. The PIC and PICS are the methods from the paper~\cite{lei2023text+}. The PerCo\textsubscript{19} and the PerCo\textsubscript{313} represent the pre-trained PerCo model~\cite{lei2023text+} corresponding to 0.0019 bpp and 0.0313 bpp, respectively. The MS-ILLM\textsubscript{20}, the MS-ILLM\textsubscript{40}, and the MS-ILLM\textsubscript{350} represent the pre-trained MS-ILLM model corresponding to 0.0020 bpp, 0.0040 bpp, and 0.0350 bpp, respectively.

\noindent \textbf{(1) Depth estimation.} We evaluate the performance of depth estimation for the reconstructed images using the pre-trained Depth-Anything-V2-Large model of Depth Anything V2~\cite{yang2024depth} on the KITTI~\cite{Uhrig2017THREEDV} depth validation set with the size $1216 \times 352$. To demonstrate the generalization ability of our methods, we also test on an indoor scene dataset, that is Hypersim~\cite{roberts:2021}. For evaluation metrics, $\delta_i = \text{percentage of max} \left ( d^* / d \right ) < 1.25^i$, where $i = 1, 2, 3$, and $d^*$ is the model prediction result and $d$ is the ground truth. ``AbsRel" represents the absolute relative error, given by $|d^* - d| / d$. ``RMSE" is the root mean square error between the model prediction and the ground truth. ``RMSE log" is the root mean square error of logarithms. As shown in Table~\ref{tab: depth}, our method with $C=1$ uses only 0.0069 bpp but achieves better performance than PICS using 0.0235 bpp and MS-ILLM\textsubscript{350} using 0.0539 bpp in the KITTI dataset. Furthermore, our method with $C=2$ using 0.0074 bpp outperforms the PerCo\textsubscript{313} using 0.0329 bpp. On the Hypersim dataset, our method with $C=8$ uses only 0.0043 bpp but achieves better performance than other methods except PerCo\textsubscript{313}. And our methods with $C=16$ use 0.0064 bpp to achieve better performance than Perco\textsubscript{313} using 0.0313 bpp.

\begin{table}
    \centering
    \caption{Object detection result on the subset of COCO2017.}
    \label{tab: obj. detect}
\resizebox{\linewidth}{!}{
    \begin{tabular}{lccccccc}
    \toprule
    Method & bpp$\downarrow$ & P$\uparrow$ & R$\uparrow$ & mAP50-95$\uparrow$ & mAP50$\uparrow$ & mAP75$\uparrow$\\
    \midrule
    Original & - & 0.944 & 0.743 & 0.763 & 0.835 & 0.799 \\
    Directly Gen.  & 0.0037 & 0.481 & 0.350 & 0.092 & 0.327 & 0.022 \\
    PIC & \textbf{0.0027} & 0.404 & 0.149 & 0.053 & 0.125 & 0.026 \\
    PICS  & 0.0221 & 0.815 & 0.656 & 0.584 & 0.708 & 0.631 \\
    PerCo\textsubscript{19} & 0.0037 & 0.813 & 0.673 & 0.455 & 0.709 & 0.507 \\
    PerCo\textsubscript{313} & 0.0329 & 0.900 & 0.720 & 0.697 & 0.785 & 0.760 \\
    MS-ILLM\textsubscript{20} & 0.0086 & 0.405 & 0.217 & 0.123 & 0.179 & 0.132 \\
    MS-ILLM\textsubscript{40} & 0.0122 & 0.406 & 0.350 & 0.273 & 0.344 & 0.274 \\
    MS-ILLM\textsubscript{350} & 0.0496 & 0.857 & 0.536 & 0.600 & 0.679 & 0.635 \\
    \midrule
    Ours ($C=1$) & 0.0044 & 0.894 & 0.699 & 0.568 & 0.767 & 0.601  \\
    Ours ($C=2$) & 0.0051 & 0.838 & 0.710 & 0.623 & 0.765 & 0.685 \\
    Ours ($C=4$) & 0.0063 & 0.903 & \textbf{0.752} & 0.697 & 0.813 & 0.774 \\
    Ours ($C=8$) & 0.0094 & 0.882 & 0.739 & 0.701 & 0.817 & 0.746 \\
    Ours ($C=16$) & 0.0155 & \textbf{0.938} & 0.733 & \textbf{0.744} & \textbf{0.820}& \textbf{0.781} \\
    \bottomrule
    \end{tabular}
    }
\end{table}

\begin{figure*}[h]
    \centering
    \includegraphics[width=0.9\linewidth]{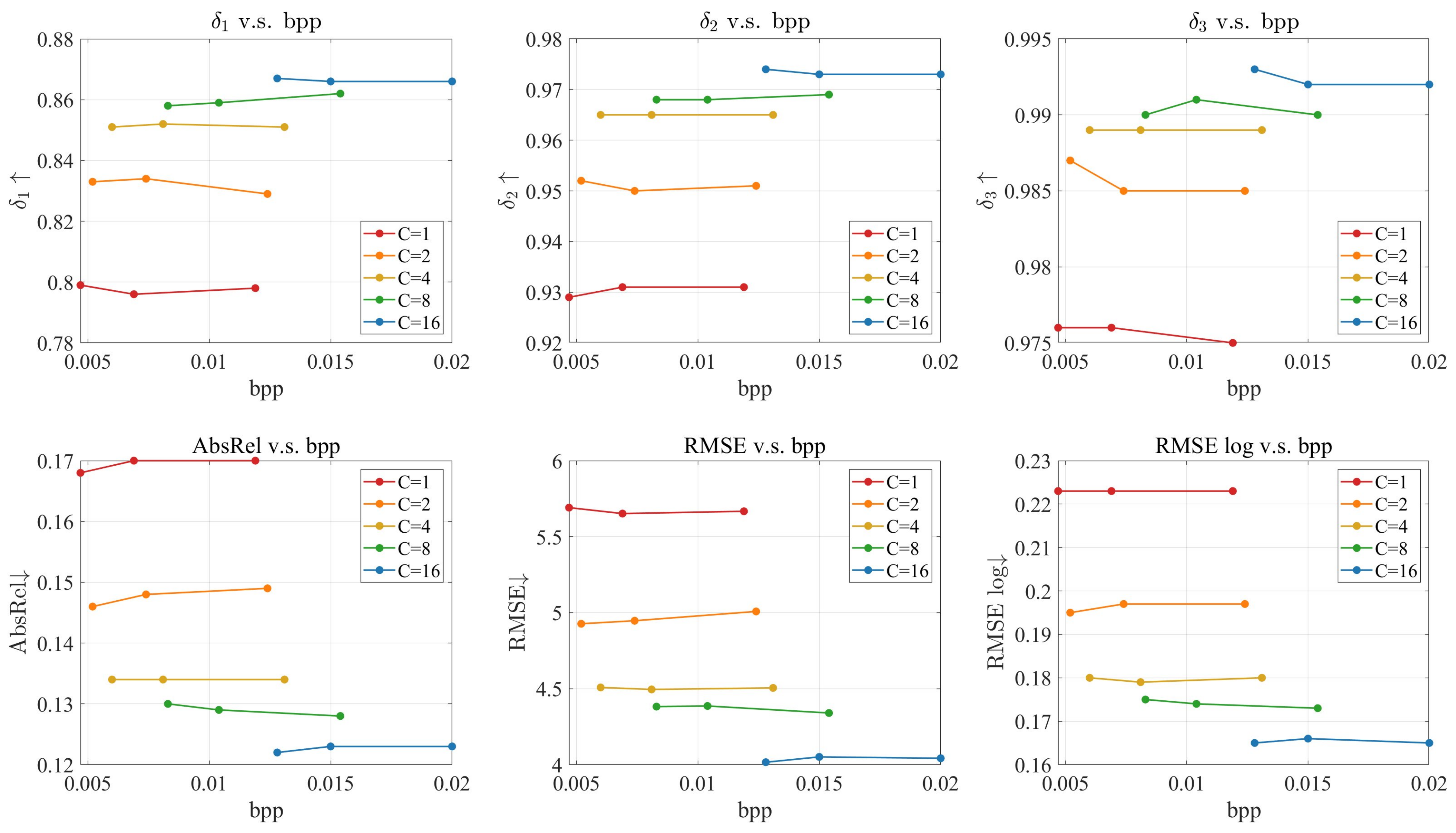}
    \caption{Ablations of the prompt with different lengths on the reconstruction quality in the depth estimation task using the KITTI sub test set.}
    \label{fig:j ablation of prompts}
\end{figure*}

\noindent \textbf{(2) Semantic segmentation.}
We conduct semantic segmentation experiments on the Cityscapes~\cite{cordts2016cityscapes} semantic segmentation validation set with the size $2048 \times 1024$, using the Mask2Former~\cite{cheng2021mask2former} of open-mmsegmentation~\cite{mmseg2020} with backbone Swin-L (in 22k). As shown in Table~\ref{tab: pix-semantic-seg}, our method with $C=8$ uses only 0.0023 bpp, outperforming all the other methods with much higher bit rates. Although the PIC uses only 0.0020 bpp, its results are even worse than the results directly generated by FLUX~\cite{flux2024}, and our method with $C=1$ uses only 0.0011 bpp to have better results. Figure~\ref{fig: qualitative results} also demonstrates our method's superior performance in preserving detailed structural information.

\noindent \textbf{(3) Object detection.}
We evaluate object detection with the pre-trained YOLO11x of Ultralytics~\cite {Jocher_Ultralytics_YOLO_2023} on the COCO2017~\cite{lin2014microsoft} validation set. As shown in the Table.~\ref{tab: obj. detect}, our method with $C=4$ uses less bpp but achieves better performance than other methods. Although PIC uses the least bpp among all methods, its performance is even worse than the performance of results directly generated by FLUX~\cite{flux2024}.

\noindent \textbf{(4) Comparison on traditional compression performance.} Our method not only performs very well on vision task-oriented image compression, but also achieves superior performance in conventional image compression.  We conduct experiments on the Kodak~\cite{kodak_dataset}. Figure~\ref{fig: rd curve of Kodak} shows the results of PSNR, MS-SSIM~\cite{wang2003multiscale}, and LPIPS~\cite{zhang2018unreasonable}, and our method achieves the best performance among all of them.

\begin{figure}[h]
    \centering
    \includegraphics[width=\linewidth]{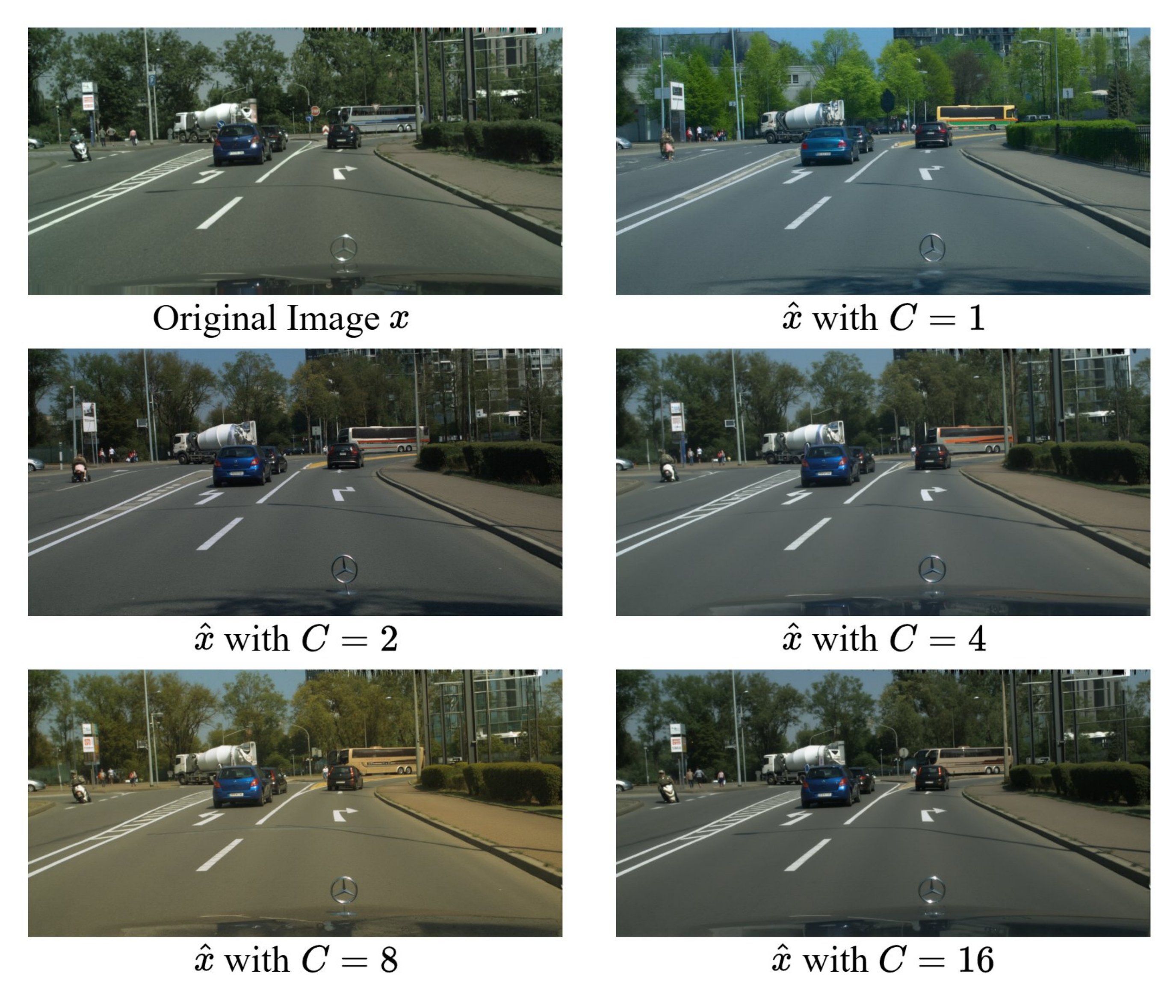}
    \caption{Reconstructed images with different numbers of channels.}
    \label{fig: ablation visualization}
\end{figure}

\subsection{Ablation Studies}
In the following, we provide detailed ablation studies to further understand our proposed method. 

\noindent \textbf{(1) Ablation studies on the number of channels.}
We change the selected structural guidance latent $\hat{y}_{sel}$ in 1, 2, 4, 8, 16 to examine the impact on compression performance. Figure~\ref{fig: ablation visualization} shows that the reconstructed image has more details aligning with the original one as more channels. Table~\ref{tab: depth},~\ref{tab: pix-semantic-seg} and~\ref{tab: obj. detect} also shows that more channels in the structural guidance latent usually lead to better performance, as more information has been used in the generation. However, these channels contain redundant and noisy information, so more channels don't always perform better than fewer channels, and adding more channels on top of one channel does not significantly improve the effect. 

\noindent \textbf{(2) Ablation studies on the length of the prompt.} 
We conduct experiments with different lengths of prompts to evaluate the effect of prompts and the robustness of our method. Figure~\ref{fig:j ablation of prompts} shows the reconstruction quality in the depth estimation task on the KITTI sub test set. As longer prompts are used, higher bpp would be in the same number of channels, and it will give more detailed information about images. However, results show that our method achieves similar results using different lengths of the prompt, which means our method has great robustness.

In the Supplemental Materials, we have provided more experimental results to demonstrate the superior performance of our proposed GSC method.

\section{Discussion}
Semantic communication aims to interpret information at the semantic level and transmit representations that accurately convey the intended meaning, which is similar to the task in this paper. However, existing methods designed for semantic communication~\cite{TransTIC, li2024image, zhang2024allinone} primarily target bitrates above 0.1 bpp, making them unsuitable for the extremely low-bitrate scenarios considered in this paper.

\section{Conclusion}
\label{sec: conclusion}
We have developed Generative Semantic Coding (GSC), a new deep learning-based image compression method that uses multiple latent channels to guide the generation of images that preserve structural information as the original images while using less than 0.007 bpp. We developed new methods for constructing structural guidance and effectively utilizing it during the image generation process. This method will be very useful in scenarios where the communication channel conditions are very challenging and the bandwidth is very limited, however, both the sender and receiver have sufficient computational resources.
Theoretical analysis is conducted to determine the lower bound of the compression. Future work includes eliminating redundant and noisy information in the latents to enhance compression and achieve a flexible balance between compression efficiency and visual analysis quality.

\section*{Acknowledgments}

This work is supported by the National Natural Science Foundation of China (No. 62331014) and Project 2021JC02X103. We acknowledge the computational support of the Center for Computational Science and Engineering at Southern University of Science and Technology.

{
\small

}
\end{document}